\documentclass{article}
    \usepackage{arxiv}
    \usepackage{moreverb,url}
    \usepackage[hidelinks]{hyperref}
    \usepackage{IEEEtrantools}
    \usepackage[square,numbers]{natbib}
    \usepackage[pdftex]{graphicx}

\usepackage{amsmath}
\usepackage{array}
\usepackage{booktabs}

\usepackage[caption=false,font=footnotesize]{subfig}

\usepackage{dblfloatfix}

\usepackage{url}

    \newcommand{\resizeboxM}[3]{#3}
    \def\imwidth{73mm}
    \def\widewidth{100mm}

\def\eg{e.g.,\ }

\usepackage{xcolor}
\usepackage[normalem]{ulem}
\newcommand\soutpars[1]{\let\helpcmd\sout\parhelp#1\par\relax\relax}
\long\def\parhelp#1\par#2\relax{%
  \helpcmd{#1}\ifx\relax#2\else\par\parhelp#2\relax\fi%
 }

\newcommand{\Argmin}[1]{\ensuremath{\mathrm{Arg}\underset{#1}{\mathrm{min}\,}}}

\newcommand{\Source}{\ensuremath{\mathrm{Source}}}
\newcommand{\Target}{\ensuremath{\mathrm{Target}}}

\newcommand{\titleName}{
Unsupervised Transfer Learning for Anomaly Detection: Application to Complementary Operating Condition Transfer
}

\newcommand{\abstr}{
Anomaly Detectors are trained on healthy operating condition data and raise an alarm when the measured samples deviate from the training data distribution. This means that the samples used to train the model should be sufficient in quantity and representative of the healthy operating conditions. But for industrial systems subject to changing operating conditions, acquiring such comprehensive sets of samples requires a long collection period and delay the point at which the anomaly detector can be trained and put in operation.

A solution to this problem is to perform unsupervised transfer learning (UTL), to transfer complementary data between different units. In the literature however, UTL aims at finding common structure between the datasets, to perform clustering or dimensionality reduction. Yet, the task of transferring and combining complementary training data has not been studied.

Our proposed framework is designed to transfer complementary operating conditions between different units in a completely unsupervised way to train more robust anomaly detectors. It differs, thereby, from other unsupervised transfer learning works as it focuses on a one-class classification problem. The proposed methodology enables to detect anomalies in operating conditions only experienced by other units. The proposed end-to-end framework uses adversarial deep learning to ensure alignment of the different units' distributions. The framework introduces a new loss, inspired by a dimensionality reduction tool, to enforce the conservation of the inherent variability of each dataset, and uses state-of-the art once-class approach to detect anomalies. We demonstrate the benefit of the proposed framework using three open source datasets.
}

\newcommand{\keywo}{Anomaly Detection, Fault detection, Unsupervised Domain Adaptation, Aircraft, Bearings, Image Processing.}

\newcommand{\aknow}{This work was supported by the Swiss National Science Foundation (SNSF) Grant no. PP00P2-176878.}

\begin{document}

    \title{\titleName}
    \author{%
        Gabriel Michau\\
        ETH Z\"urich,\\
        Z\"urich, Switzerland\\
        \And 
        Olga Fink\\
        ETH Z\"urich,\\
        Z\"urich, Switzerland}
        \subtitle{Preprint}
        \date{24th of November 2020}
        \maketitle
        \begin{abstract}
        \abstr
        \end{abstract}
        \keywords{\keywo}

\section{Introduction}

\begin{figure}
\centering
\includegraphics[width=\widewidth]{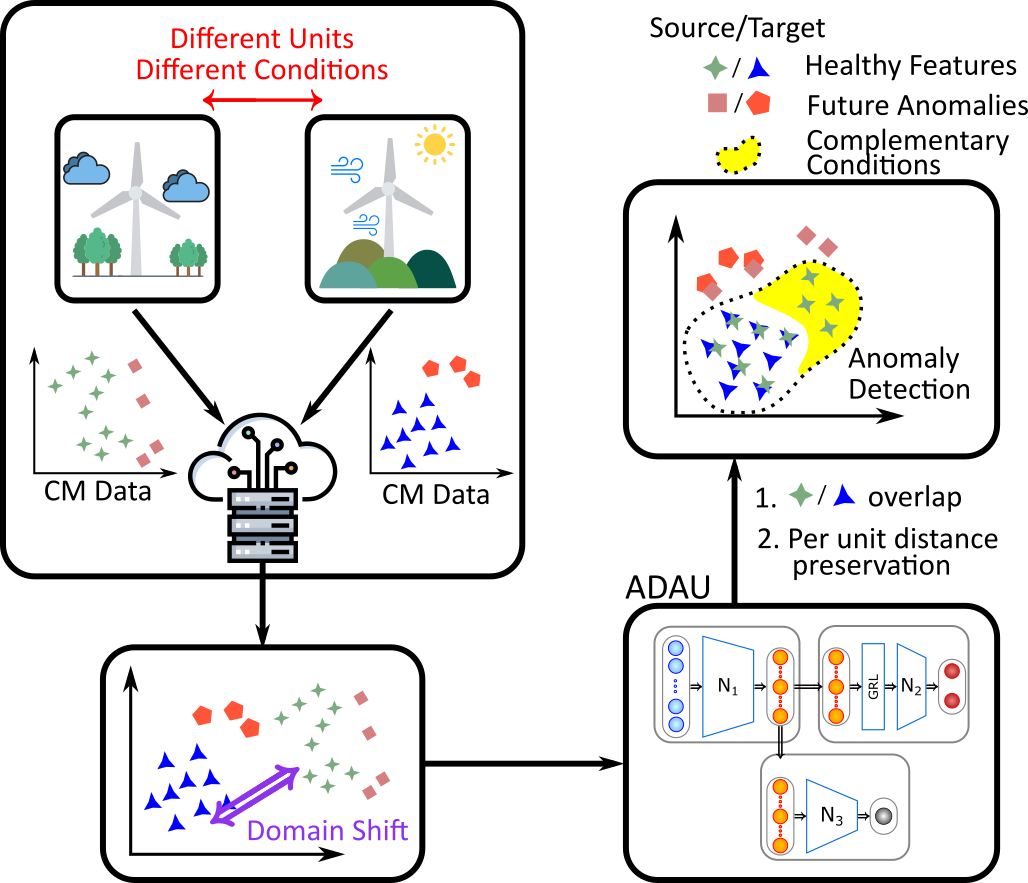}
\caption{\textbf{Domain Adaptation for Anomaly Detection}. Condition Monitoring (CM) data are collected on each unit from a fleet. Due to each unit's characteristics and environmental conditions, the data belong to different distributions, a phenomenon known as ``Domain Shift''. Alignment is performed with the proposed ADAU approach using healthy CM data only (blue triangles and green stars). The final monitoring system is able to detect anomalies.}
\label{fig:ga}
\end{figure}

Traditional machine learning relies on the assumption that the training and the test data are drawn form the same distribution. This assumption implies that any change in the test distribution renders the trained model obsolete, and requires the development of a new model with data collected from the new distribution or at least an update of the trained model. 
This shortcoming of machine learning is particularly problematic for industrial applications. Operators are usually monitoring fleets of units, each with their own particularities. Different hardware and software versions, different lifetimes and different environmental and operational conditions usually hinder the development of machine learning models valid across all units. The alternative, consisting of training one model per unit, is time consuming and requires large amounts of data for every unit. Thus, it is not always possible, \eg for newly installed units, or for systems subject to evolving operating or environmental conditions~\cite{Michau2018b}. For such cases, it would be beneficial to be able to \textbf{transfer} information between units and between models, such as to be able to train models that are robust to newly encountered conditions. This is particularly crucial for units with only little or non-representative training data.

``Transfer learning'' encompasses all the techniques aiming at minimising the efforts in the development of a new model, by transferring knowledge or information gathered on the source distribution to the model handling the target distribution ~\cite{Pan2010}. Among the transfer learning tasks, one distinguishes, first, ``Inductive'' transfer learning, when the distribution change is in the output's space, due to the necessity of solving a different machine learning task (\eg a classification with more, less or different classes). Second, ``Transductive'' transfer learning tackles the cases of models trained in a supervised manner with the source data that need to handle a distribution change in the input's space (data collected in different conditions, with different sensors, on different systems) and where the aim is to solve the same machine learning task. Last, ``Unsupervised'' transfer learning focuses on unsupervised machine learning tasks in both the source and the target domains. Since the tasks are unsupervised, the distribution changes to handle are in the input space and it is not relevant whether the same task is solved for both source and target or not.

For industrial applications, in particular for the monitoring of complex industrial systems, the machine learning tasks are often unsupervised. Complex industrial systems have a large number of parts and components interacting with each other, making the number of possible faults uncountable. Since these systems are also reliable by design, collecting a labeled dataset with enough instances of each possible fault is unrealistic. Therefore, the monitoring is often tackled as an anomaly detection task~\cite{michau2020feature}, whereby a model is trained to recognise data from healthy operating conditions only, raising an alarm when the test data are sufficiently different from those seen in training.

The literature contains a large body of contributions on inductive and transductive transfer learning since they are the natural extensions of supervised learning problems: once a lot of effort has been made to develop a model on a supervised, clean and controlled dataset, transfer learning aims at maximising the return on investment by making the model applicable in as many real-life applications as possible. Research on \textbf{unsupervised transfer learning}, however, has been more limited. Most works focus on the machine learning tasks of clustering and of dimensionality reduction~\cite{Pan2010,tan2018survey,redko2019advances}. The anomaly detection task is quite different from clustering or dimensionality reduction that assume an underlying common structure in the data (\eg that somehow both datasets could be clustered with the same clusters, or for dimensionality reduction, that there are common factors of variability in both datasets). On the contrary, transfer learning for anomaly detection aims at transferring complementary data of the main (and only) class and to learn the combination of the complementary operating conditions, rather than their commonalities, while mitigating the distribution shift due to the datasets' particularities. 

In this paper, we propose a solution for unsupervised transfer learning for anomaly detection, a machine learning task not yet tackled in the literature to the best of our knowledge. The task at hand, anomaly detection, is unsupervised with respect to the anomalies and is by definition an \emph{Unsupervised Transfer Learning} task. We approach this task as an unsupervised one-class classification problem, that is, we assume that only data belonging to the main (and only) class are available at training time. Thus, the transfer learning task at hand has some common points with traditional transductive learning, and more specifically with domain adaptation: handling a distribution change in the input space to solve the same machine learning task of anomaly detection. In our case however, the task is unsupervised with respect to the anomalies. Our proposed framework, illustrated in Figure~\ref{fig:ga} combines adversarial deep learning, that has brought state-of-the-art results in transductive transfer learning~\cite{ganin2016domain,wang2019domain,wang2020missing}, with dimensionality reduction inspired concepts, frequently used in unsupervised transfer learning~\cite{Pan2010,sanodiya2019novel}. In fact, transferring the complementary operating conditions results in extending the operating conditions of the target unit and enlarging the representativeness of the dataset. 

We test the proposed approach on three open datasets of varied nature. First, we apply the methodology to vibration data from bearings under different loads, the Case Western Reserve University bearing dataset~\cite{smith2015rolling}. The anomaly detection task is the detection of any faults present in the dataset and the domain adaptation aims to perform the adaptation from one load to another. Second, we apply the methodology to the monitoring of turbofan engines in operation, simulated using the AGTF-30 model~\cite{chapman2017control} with real flight profiles. In this case, the anomaly detection task is the detection of performance degradation in the fan, as well as in the high- and low-pressure turbines efficiency and capacity. The domain adaptation task is the alignment between two flights with different engine characteristics and different profiles.
Last, to show the generalisability of the proposed approach, we demonstrate that it also provides benefits for anomaly detection in a computer vision task: the handwritten digit MNIST to MNIST-M domain adaptation task~\cite{lecun1998gradient,arbelaez2010contour}. For the anomaly detection setup, one digit is regarded as the main class and all other digits as anomalies.

The remainder of this paper is organised as follows: Section~\ref{sec:RW} presents relevant works found in the literature. Section~\ref{sec:M} details the architecture used in this work and the new proposed loss. Last, Section~\ref{sec:E} presents the three experiments and their results.

\section{Related Work}
\label{sec:RW}
A distribution change in the input space of a machine learning model, due to changing conditions or different data collection setups, is denoted in the literature as a ``domain shift''. Finding models that are robust to such shifts or are able to reconcile the distributions in an intermediate latent space, also denoted as ``domain alignment'' or ``domain adaptation'' techniques, has been the focus of numerous works in the literature on transductive transfer learning~\cite{tan2018survey, redko2019advances}. It has been tackled in several ways. For example, one can first find good features for solving the machine learning task on the source dataset and then find the right transformation of the target dataset~\cite{fernando2013unsupervised, Xie2016, Zhang2017a}. The problem can also be tackled in a combined approach, such as to find features that minimise the supervised task's loss on the source dataset while maximising the overlap between source and target features~\cite{ben2012hardness}. Finding the right feature space that satisfies both conditions is the main challenge. It has also been approached with fuzzy systems~\cite{zuo2018fuzzy, xu2019transfer, xie2018generalized} (and references therein) or with deep learning~\cite{ajakan2014domain, ganin2016domain}.

Recently, deep learning progressed significantly on the domain adaptation problem, in particular due to the success of adversarial architectures~\cite{ganin2016domain,wang2019domain}. These architectures usually consist in the optimisation of a neural network in three parts: a feature encoder performing the data transformation; a feature processor, trained to solve the machine learning task of interest on the source domain (in most cases a classifier); and a feature discriminator, trained to distinguish whether the data originate from the source or target dataset. The min-max problem is solved at the level of the feature extractor. It is trained at the same time to minimise the loss of the feature processor and to maximise the discrimination loss of the discriminator. In other words, once trained, the feature extractor should provide features that are relevant for the machine learning task at hand and indistinguishable between source and target. Adversarial architectures have been implemented in several studies, performing well on the machine learning task tested on the target domain~\cite{tan2018survey}. Some works attempted to make this approach robust to changes in the output distributions (\eg assuming missing class in the source domain~\cite{wang2019domain,wang2020missing}). These approaches fail when the distribution discrepancy increases too much, since the supervised task is used to learn a relevant transformation of the data in the feature space~\cite{wang2020missing}.

In the literature, unsupervised transfer learning  usually assumes the existence a common structure between source and target and leverages this information to perform the transfer. This structure can be the knowledge of: pairwise correspondence between samples from the  source and the target~\cite{wang2008manifold}, of common clusters~\cite{dai2008self,sanodiya2020unsupervised}, of common relevant scale for spectral decomposition~\cite{chang2017unsupervised} used for fine tuning in a subsequent step, or of existing common factors of variability when performing transferred dimensionality reduction~\cite{wang2008transferred,du2013unsupervised}.
Yet, to the best of our knowledge, the unsupervised transfer of complementary samples with domain shift applied to anomaly detection, without \textit{a priori} knowledge of existing clusters has not yet been tackled in the literature.

In the context of industrial systems, this problem is particularly important for monitoring of fleets of systems originating from the same manufacturer, and thus monitored by similar sets of sensors~\cite{Leone2017}, but experiencing a domain shift due to differences in system configuration, usage, or environment.
In such context, previous works have devised ways to find units within a fleet with similar operating conditions~\cite{liu2018cyber} or with smallest domain shift ~\cite{al2018framework,Michau2018b}; yet these approaches severely restrict which units can be combined, \textit{de facto} limiting their application to cases with large fleets of units. Other works have focused on developing models robust to domain shifts, but always as a transductive transfer learning for domain adaptation. It proved useful in several contexts such as fleet monitoring~\cite{michau2019domain,moradi2020application} and simulated-to-real system adaptation~\cite{xu2019digital}. Domain adaptation has been mostly used for fault diagnostics~\cite{wang2019domain, wang2020missing, xu2019digital, jiao2019classifier, han2020deep, singh2020deep} and remaining useful life estimation~\cite{zhang2018transfer,da2020remaining}, yet it has not been studied in the context of unsupervised fault detection, apart from the preliminary studies of this work~\cite{michau2019domain}.

\section{Method}
\label{sec:M}

    \subsection{The Network}

\begin{figure}
\centering
\includegraphics[width=\imwidth]{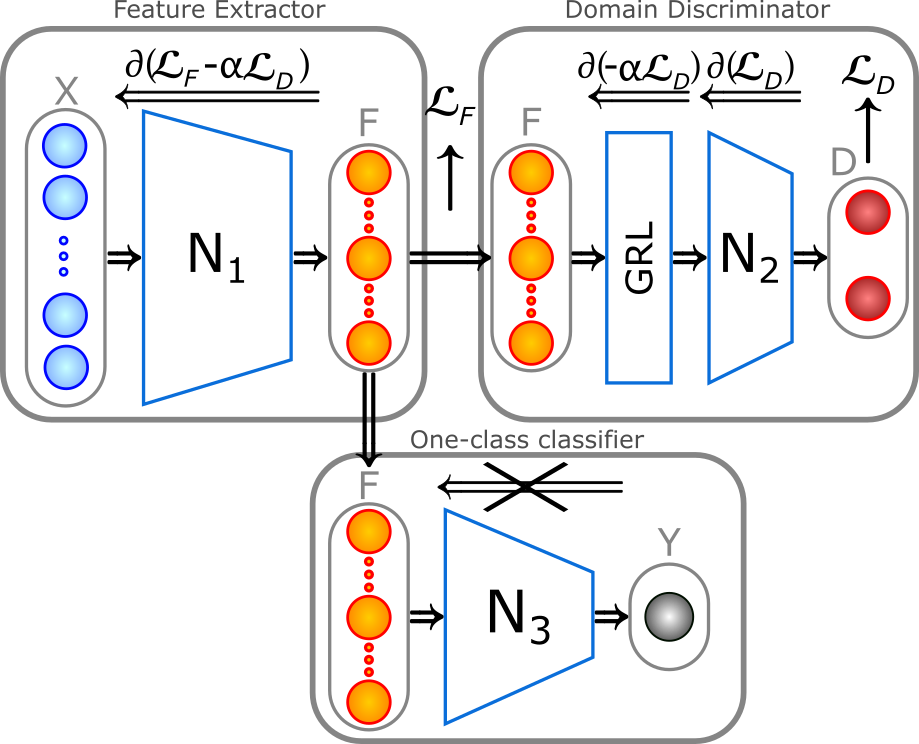}
\caption{\textbf{Adversarial Domain Adaptation for Unsupervised Anomaly Detection (ADAU)}. The feature extractor $N_1$ is trained to minimise the multidimensional scaling loss $\mathcal{L}_F$ and to maximise the domain discriminator loss $L_D$. The domain discriminator $N-2$ is a traditional feed-forward dense classifier. Once trained, features are fed to the one-class classifier for anomaly detection.}
\label{fig:nn}
\end{figure}

In this paper, we propose the use of an adversarial deep learning architecture in three parts, as illustrated in Figure~\ref{fig:nn}~\cite{ganin2016domain,wang2019domain,wang2020missing}. It consists of a feature encoder $N_1$, an adversarial feature domain discriminator connected to the network $N_1$ with a Gradient Reversal Layer~\cite{ganin2016domain}, and a feature processor $N_3$ performing machine learning tasks of interest (here the anomaly detection). 
Therefore, instead of a traditional classifier, we use a one-class classifier as feature processor $N_3$. The focus of the paper is on the alignment strategy, and on its impact on the anomaly detector performance. We perform the anomaly detection using Extreme Learning Machines since it has proven to be a simple yet efficient approach~\cite{michau2020feature} that can be easily integrated within a deep learning architecture.

Usually in DA, source labels are used to ensure that the extracted features from $N_1$ are meaningful for the task at hand~\cite{wang2020missing}, most often a classification task. In such setups, the feature extractor is trained to minimise the classification loss for samples with labels while simultaneously maximising the domain discriminator loss between source and target domain (usually a binary cross-entropy loss). Minimising the classification loss ensures that the labels act as anchors during the transformation of the data, and that the features retain discriminant characteristics for the feature processor. At the same time, features that maximise the domain discriminator loss cannot be classified by their domain of origin and are, therefore, domain independent.

For the purpose of anomaly detection as defined in our work, however, the task is unsupervised: the aim is to detect anomalies examples of which are not available at training time, neither in the source nor in the target domain. We assume, however, that a healthy set is available at the time of training for both the source and the target, albeit with very few samples for the target. The objective, therefore, is to devise domain-independent features that enable to monitor the health of the target data, including during operating modes not seen during training. 

To perform the feature alignment without labels, we seek to ensure that the features are meaningful with respect to the information the data originally contained. Therefore, we propose to minimise a loss inspired by a dimension reduction tool, multidimensional scaling~\cite{cox2001multidimensional}. By doing so, deep learning adversarial architectures, as described above, can be used. Here, the feature extractor $N_1$ can then be trained in an adversarial manner, first, to minimise the proposed multidimensional scaling loss, and second, to maximise the domain discrimination loss. Once trained, the features can finally be used for one-class classification.

    \subsection{The Multidimensional Scaling Loss}
        
Denoting the input data as $X$ and the features learned by the neural network as $F$, we propose to define the multidimensional scaling loss as
\begin{equation}
\resizeboxM{0.95\columnwidth}{!}{%
\mathcal{L}_F = \sum_{S\in \left\lbrace \substack{\Source\\\Target} \right\rbrace} \frac{1}{\vert S \vert}  \sum_{(i,j)\in S}\left\Vert \left\Vert X_i - X_j \right\Vert_2 - \widehat{\eta}_S \left\Vert F_i - F_j \right\Vert_2 \right\Vert_2%
},
\label{eq:MDSLoss}
\end{equation}
where
\begin{equation}
\forall S\in \left\lbrace \substack{\Source\\\Target} \right\rbrace, \quad \widehat{\eta}_S = \Argmin{\tilde{\eta_S}} \mathcal{L}_F (\tilde{\eta}_S).
\end{equation}

From an optimisation perspective, the minimisation of this loss is equivalent to that where $\eta_{\Source}=1$ and $\eta_{\Target}= \widehat{\eta}_{\Source}/ \widehat{\eta}_{\Target}$ such that we can define the loss as
\begin{equation}
\resizeboxM{0.95\columnwidth}{!}{%
\mathcal{L}_F = \sum_{S\in \left\lbrace \substack{\Source\\\Target} \right\rbrace} \frac{1}{\vert S \vert}  \sum_{(i,j)\in S}\left\Vert \left\Vert X_i - X_j \right\Vert_2 - \eta_S \left\Vert F_i - F_j \right\Vert_2 \right\Vert_2%
},
\end{equation}
where
\begin{equation}
\eta_{\Source}=1,\quad \eta_{\Target} = \Argmin{\tilde{\eta}_\Target} \mathcal{L}_F (\tilde{\eta}_{\Target}).
\label{eq:argmin}
\end{equation}
Equation~\eqref{eq:argmin} has a closed-form solution that allows one to compute at each training step the optimal scaling parameter given the current features. Since the scaling parameter of the source dataset, $\eta_{\Source}$, is fixed to $1$, and since source and target will be encouraged to overlap to maximise the domain discriminator loss, the values for $\eta_{\Target}$ are in fact quite constrained and did not lead to instabilities during training.

The minimisation of this loss encourages the latent space to conserve inter-sample relationships for both the source and target sets. The latent space is a non-linear transformation of the original space such that the between-sample distances are preserved as well as possible. Our proposed loss considers source and target independently, thus allowing for independent scaling and possibly mitigating distribution shifts due to translation, noise, rotations, and scales. The adversarial discriminator will also ensure that both source and target distributions overlap in the latent space.

\section{Experiments and Results}
\label{sec:E}

    \subsection{Experimental Design}
To demonstrate the effectiveness of the approach, we test the proposed architecture and loss on three open datasets. The experimental design in similar in the three cases. First, the anomaly detection is performed with Hierarchical Extreme Learning Machines (HELM)~\cite{michau2020feature} on the target domain with a decreasing number of samples used for the training. The number of training samples is reduced either until it becomes too small to train a neural network, or until the  balanced accuracy (BA) of the detection shows a clear drop. Second, this insufficient number of samples is used as the target training set and another dataset with different conditions is used as the source. These two datasets are then used to train, on the one side, the proposed architecture (ADAU) and, on the other side, another HELM as a baseline. 

\textbf{Architecture.} %
Finding optimal network architectures for unsupervised tasks is an open research question, one not yet solved. Due to the impossibility of validating the model on the basis of its anomaly detection abilities--since it is assumed that, at training time, no examples of faults are available--the traditional training/validation split or cross-validation cannot be performed. Thus we designed the networks based on our experience with HELM used as an anomaly detector. HELM consists in the stacking of a single-layer auto-encoder, with a single-layer one-class classifier ELM trained on the target value $1$. In short, the auto-encoder devises features that are combined in the second layer into a single scalar, whose distance to the training target value is monitored. For each sample, this value is compared to a statistical descriptor of this distance computed with a healthy validation dataset. Here, this descriptor is 1.2 times the 99.5th percentile, as in~\cite{michau2020feature}. The validation set is always set to a size of 25\%, that of the training set size, mimicking an 80\%/20\% split.
Empirically, the number of neurons used for HELM can be set according to the elbow approach~\cite{thorndike1953belongs} applied to the reconstruction loss for the auto-encoder, and to the distance to the target value $1$ for the subsequent one-class classifier.
Once these numbers have been established, the feature extractor $N_1$ is designed as a two-layer feed-forward dense network with the same neuron number as the HELM auto-encoder. The one-class classifier $N_3$ has the same number of neurons as the HELM one-class classifier, and the domain discriminator is a dense network with two 5-neuron layers. The gradient reversal layer hyper-parameter $\alpha$, weighting the back propagated domain discrimination loss gradient to the feature extractor parameters, is set to $0.1$. It has been shown in the literature that a successful adversarial training requires particular focus on the training of the domain discriminator~\cite{ganin2016domain}.
This approach has been chosen for its simplicity and because it yielded consistent results between datasets and experiments, without noticeable training instabilities.

\textbf{Training.} %
All neural networks, except the ELMs, are trained using Tensorflow with the Adam optimiser at a learning rate of $10^{-3}$, over $2\,000$ epochs, a sufficiently large number to observe the convergence of the different losses at the end of the training.

\textbf{Metrics.} %
In all experiments, we choose to report for the target the false positive rate (FPR) and the balanced accuracy (BA), the latter defined as the average between the true positive rate (TPR) and the false positive rate. 

The choice of the balanced accuracy as a metric does not require any assumption concerning the ratio of faulty versus healthy samples in the dataset. It is also linear with respect to the TPR and the FPR, making the changes in its value easily interpretable. %
Also, since we report the BA, the FPR and the number of samples tested, TP, TN, FP and FN can be recovered for the computation of any other metric of choice, such as the $F_1$-score, the Matthews correlation coefficient, including in different unbalanced scenarios, under the common assumption of model's constant specificity (TPR) and sensitivity (1-FPR). From an application perspective, we believe FPR and TPR need to be jointly evaluated, FPR to assess whether the rate of false alarms is acceptable and TPR to decide whether the model is worth implementing. The right values are therefore application-dependent. One can also note that the TPR and FPR values are linked via the detection threshold used in the one-class classification. Lowering the threshold will mechanically improve the TPR at the expense of the FPR and vice versa.

\textbf{Repetition of the experiments and Visualization of the Results.} %
Every experiment is repeated five times, such that we also report the standard deviation for each indicator. In all figures, the BA for the HELM trained on the available healthy target data is plotted in blue, the BA for the HELM trained on the combined source and target data is plotted in yellow, and the BA for ADAU is plotted in green. For each result, the corresponding number of false positives, when strictly positive, is plotted in red. Each bar is associated with an error bar representing the standard deviation of the indicator.

\textbf{Statistical significance.} %
For each model, we test the hypothesis that its output is not statistically different from other models outputs. This hypothesis is tested in two ways: first by performing a logistic regression on the models' success (TP and TN versus FP and FN) with a generalised linear model~\cite{Dobson2008} where the model is used as a categorical explanatory variable. The $p$-value of the variable \emph{model} allows us to accept or reject the hypothesis that the model is a factor of influence of the outputs. Second, we perform the McNemar's test~\cite{mcnemar1947note,Dietterich1998} since for each experiment, the different models are tested on the same samples. Last, these indicators are also computed by testing together all models without alignment against all models using ADAU.

    \subsection{2 HP Reliance Electric Motor Bearings Monitoring}

\textbf{Experiment.} %
The first experiment to test the proposed approach is performed on the Case Western Reserve University bearing dataset~\cite{smith2015rolling}. The dataset contains healthy and faulty condition data under three different load modes. It is therefore a dataset often used to demonstrate the performance of domain adaptation for prognostics~\cite{wang2019domain,wang2020missing,li2019multi,wang2020triplet}. Previous works have shown that the most difficult transfer task is the transfer of the model from the lowest machine load (labeled load 0 in the data) to the highest load (labeled as 3).

We therefore focus on this transfer task in the present experiment. We train the model on healthy data from load mode 0 taken as source and on load mode 3 taken as target. We then look at the aggregated detection ability over all 15 available faults. The types of faults ``Inner Race'' and ``Ball'' are tested with defect sizes of 7, 14, 21 and 28 mm; the type of fault ``Outer Race'' is tested with defect sizes of 7 and 21 mm; and the centered ``Outer Race'' is additionally tested with a defect size of 14 mm.

\textbf{Pre-processing.} %
The data pre-processing follows the setup proposed in~\cite{li2018cross}, where data at 12 kHz are used. Since the healthy data are recorded at 48 kHz, they are first down-sampled to 12 kHz. Each recording is split into 200 sequences of size 1024 using an adequate stride. On these sequences, the fast Fourier transform (FFT) is performed to obtain 512 Fourier coefficients used as inputs to the models. For example, the healthy dataset of load mode 0 has 243\,938 samples at 48 kHz, brought down to 60\,985 samples at 12 kHz, or 200 sequences of size 1024 using a stride of 300.
Similarly, for load mode 3, it has 485\,643 samples at 48 kHz or 121\,411 samples at load 12 kHz, that is 200 sequences of size 1024 using a stride of 602.

Performing the elbow method on HELM for these data, we find that HELM can be designed with 10 neurons in the auto-encoder and 50 neurons in the one-class classifier. ADAU is then constructed using these values.
\begin{figure*}
\centering
\subfloat{\includegraphics[width=\imwidth]{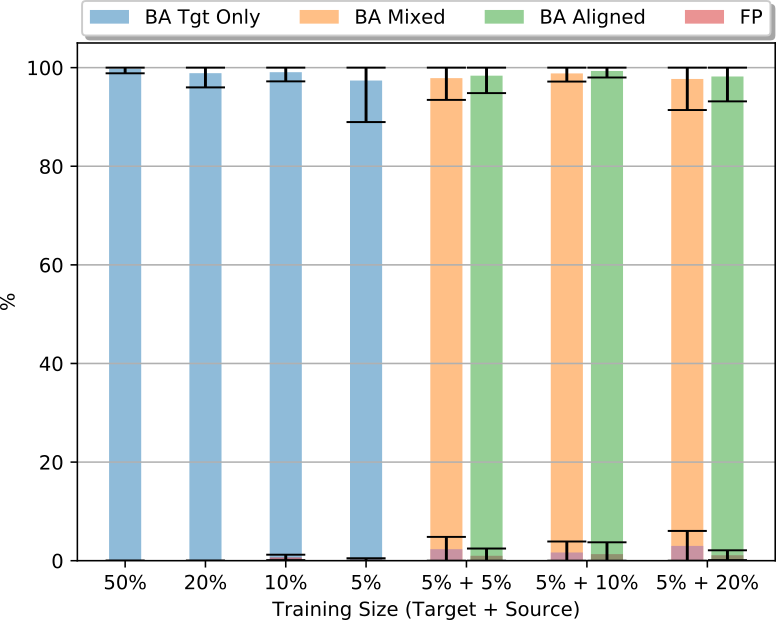}\label{sfig:CWRUa}}
\subfloat{\includegraphics[width=\imwidth]{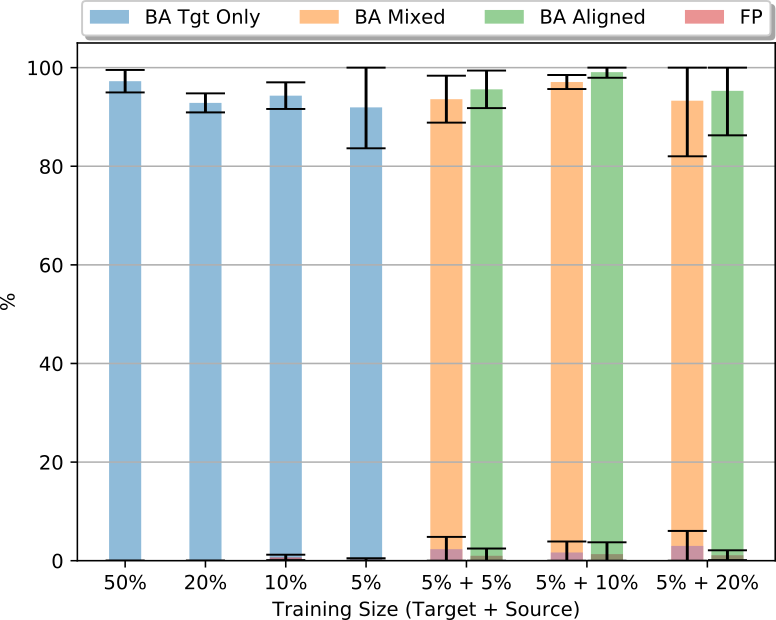}\label{sfig:CWRU14}}
\caption{\textbf{Anomaly Detection Transfer on CWRU (load 3 to 0)}. (a) All faults (b) Ball 14 mm Fault (lowest BA among all faults for HELM trained with $5\%$ of the healthy set at load 0). }
\label{fig:CWRUA}
\end{figure*}

\begingroup
\setlength{\tabcolsep}{2.9pt}
\begin{table}[]
\centering
\caption{CWRU - Balanced Accuracy (BA), False Positive rate (FP) and their standard deviation (Std) for the CWRU 0$\rightarrow$3 experiment for different training size}
\begin{tabular}{lrrrr|rr|rr|rr}
\toprule
& \multicolumn{4}{c|}{Target Only} & \multicolumn{6}{c}{5\% Target + Source}\\
& \multicolumn{4}{c|}{HELM} & \multicolumn{1}{c}{M} & \multicolumn{1}{c|}{A} & \multicolumn{1}{c}{M} & \multicolumn{1}{c|}{A} & \multicolumn{1}{c}{M} & \multicolumn{1}{c}{A}\\
& 50\% & 20\% & 10\% & 5\% & \multicolumn{2}{c|}{+ 5\%} & \multicolumn{2}{c|}{+10\%} & \multicolumn{2}{c}{+20\%}\\
\midrule
\multicolumn{11}{c}{All 15  faults}\\
\midrule
BA  & 99.77 & 98.88 & 99.08 & 97.39 & 97.87 & 98.57 & 98.83 & \textbf{99.80} & 97.70 & 98.45 \\
Std & 0.93  & 2.89  & 1.86  & 8.42  & 4.42  & 3.53  & 1.66  & 1.33  & 6.31  & 5.04  \\
FP  & 0.00  & 0.00  & 0.67  & 0.22  & 2.33  & 1.00  & 1.67  & 1.33  & 3.00  & 1.11  \\
Std & 0.00  & 0.00  & 0.54  & 0.27  & 2.49  & 1.47  & 2.22  & 2.40  & 3.03  & 0.99  \\
\midrule
\multicolumn{11}{c}{Defect on Ball, 14 mm}\\
\midrule
BA  & 97.25 & 92.85 & 94.32 & 91.94 & 93.60 & 95.60 & 97.08 & \textbf{99.08} & 93.29 & 95.29 \\
Std & 2.28  & 1.93  & 2.70  & 8.06  & 4.76  & 3.81  & 1.42  & 1.14  & 11.28 & 9.02  \\
FP  & 0.00  & 0.00  & 0.67  & 0.22  & 2.33  & 1.00  & 1.67  & 1.33  & 3.00  & 1.11  \\
Std & 0.00  & 0.00  & 0.54  & 0.27  & 2.49  & 1.47  & 2.22  & 2.40  & 3.03  & 0.99 \\ \bottomrule
\multicolumn{11}{l}{M: HELM trained with Mixed dataset (Source + Target)}\\
\multicolumn{11}{l}{A: Results using ADAU}\\
\end{tabular}
\label{tb:CWRU}
\end{table}
\endgroup

\textbf{Results.} %
The results of this experiment are reported in Table~\ref{tb:CWRU} and illustrated in Figure~\ref{fig:CWRUA}, first for the aggregated balanced accuracy over the 15 available faults and second for the fault with lowest balanced accuracy when HELM is trained with 5\% of the data, that is, the Ball defect of size 14 mm.

From the results, one can see that the bearing dataset has very little variations within each load mode. 5\% of the dataset is still enough to train a detection algorithm with over 97\% balanced accuracy. Nevertheless, using data from the source always improves the results and this improvement is always higher for our method, ADAU, as compared to simply combining the data with HELM.
Last, one can observe that adding more data from the source helps as long as both source and target data are in similar numbers (up to a factor 2). When too many source data are used, the performance on the target decreases.

All models are significantly different to each other with p-values on both the GLM and the McNemar tests below $10^{-4}$. Two exceptions are the mixed models trained with 5 and 20\% of the source ($p$-values of 0.24 and 0.26 for GLM and McNemar respectively) and the ADAU models trained with 10 and 20\% of the source ($p$-values of 0.16 and 0.20). This demonstrates the robustness of the ADAU approach to the number of added samples from the source. At the global level (all mixed models versus all ADAU), the significance is below $10^{-6}$.

    \subsection{Jet Engine Monitoring}

In the previous experiment, it appeared that if the operating mode of the target domain is stable, ADAU can offer some improvements, but they are limited due to the high performance of the anomaly detector trained solely on the target, even when trained on a very low number of samples.

\begin{figure}
\centering
\includegraphics[width=\widewidth]{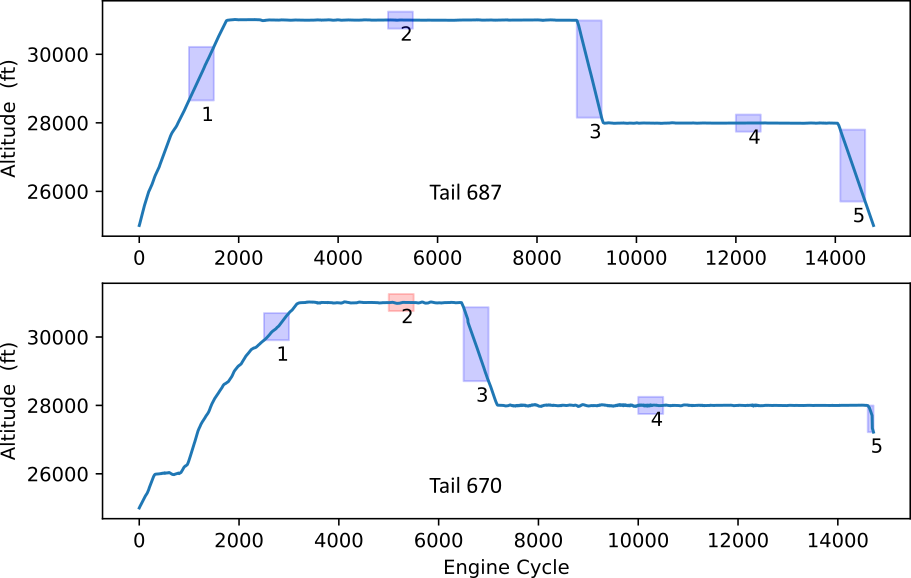}
\caption{\textbf{Altitude Profile Source and Target}. Plots of the profile altitude used for the simulation of both datasets (other simulation inputs are Mach number and power lever angle). The rectangles represent the data sampled for each mode.}
\label{fig:AGTF30Profile}
\end{figure}

\textbf{Experiment.} %
In this experiment, we propose to test the methodology on an adaptation task where source and target have five operating modes, including three transients, but only one is available for the target at training time.
For this experiment, we use the Advanced Geared Turbofan 30,000 (AGTF30)~\cite{chapman2017control} to simulate flight data for two different planes. We take as a simulation input two flight profiles (altitude, Mach number and power lever angle) downloaded from the flight tail NASA open source repository, tail 670 and tail 687\footnote{\url{https://c3.nasa.gov/dashlink/projects/85/resources/?type=ds}}. The flight profile altitudes are represented in Figure~\ref{fig:AGTF30Profile}. The source unit is tail 687, simulated with the default parameters of the AGTF30 model. The target unit is tail 670, where the capacity and efficiency of the fan, and of the low-pressure and high-pressure turbines and compressors, have been reduced to 98\%.
For the target, we simulated six independent faults, each corresponding to a drop of 0.5\% in capacity or efficiency of the fan, the high-pressure turbine, or the low-pressure turbine.

\textbf{Pre-processing.} %
For each of the five operating modes, (1) take-off, (2) high-altitude cruising, (3) descent, (4) mid-altitude cruising, and (5) landing, as well as for each fault type, we extract 500 engine cycles, illustrated as the blue and red rectangles in Figure~\ref{fig:AGTF30Profile}. For the target, however, only the data from the second operating mode, high-altitude cruising, in the healthy condition, are available at training time. This is illustrated with the red rectangle in Figure~\ref{fig:AGTF30Profile}. 

Based on these data, we use the elbow approach to estimate the number of neurons of the network and find 30 neurons for the HELM auto-encoder, and therefore for the feature extractor, and 100 neurons for the one-class classifier.

\begin{figure*}
\centering
\includegraphics[width=\textwidth]{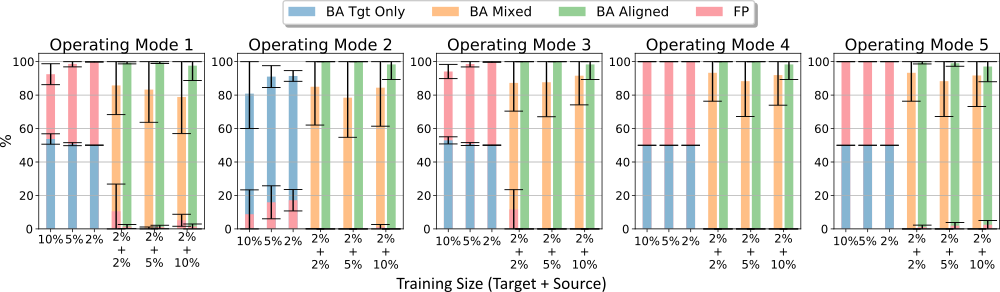}
\caption{\textbf{Anomaly Detection Transfer on Engine simulated with AGTF30.} Results for each operating mode averaged over all faults. When training with the target data of operating mode 2, most resulting models have 100\% false positive rates on other modes.}
\label{fig:AGTF30Results}
\end{figure*}

\begingroup
\setlength{\tabcolsep}{3.1pt}
\begin{table}[]
\centering
\caption{AGTF30 - Results (BA, FP, and their standard deviation Std) for the 5 operating modes using target data from mode 2.}
\begin{tabular}{lrrr|rr|rr|rr}
\toprule
& \multicolumn{3}{c|}{Target Only} & \multicolumn{6}{c}{2\% Target + Source}\\
& \multicolumn{3}{c|}{HELM} & \multicolumn{1}{c}{M} & \multicolumn{1}{c|}{A} & \multicolumn{1}{c}{M} & \multicolumn{1}{c|}{A} & \multicolumn{1}{c}{M} & \multicolumn{1}{c}{A}\\
& 10\% & 5\% & 2\% & \multicolumn{2}{c|}{+2\%} & \multicolumn{2}{c|}{+5\%} & \multicolumn{2}{c}{+10\%}\\
\midrule
\multicolumn{10}{c}{Operating mode 1: Ascent} \\
\midrule
BA  & 53.75  & 50.67  & 50.05  & 85.78 & 99.30  & 83.29 & \textbf{99.43}  & 78.89 & 97.60 \\
Std & 3.12   & 0.90   & 0.10   & 17.44 & 0.61   & 19.56 & 0.52   & 21.91 & 8.91  \\
FP  & 92.49  & 98.65  & 99.90  & 10.55 & 1.40   & 0.52  & 1.15   & 5.17  & 1.47  \\
Std & 6.24   & 1.79   & 0.21   & 16.25 & 1.23   & 0.64  & 1.03   & 3.58  & 2.29  \\
\midrule
\multicolumn{10}{c}{Operating mode 2: High-Altitude Cruising} \\
\midrule
BA  & 80.87  & 91.05  & 91.43  & 84.98 & \textbf{100.00} & 78.45 & \textbf{100.00} & 84.43 & 98.33 \\
Std & 20.86  & 6.55   & 3.20   & 22.90 & 0.00   & 23.68 & 0.00   & 23.03 & 8.98  \\
FP  & 8.78   & 15.91  & 17.15  & 0.04  & 0.00   & 0.00  & 0.00   & 1.14  & 0.00  \\
Std & 14.55  & 9.86   & 6.39   & 0.07  & 0.00   & 0.00  & 0.00   & 1.50  & 0.00  \\
\midrule
\multicolumn{10}{c}{Operating mode 3: Descent} \\
\midrule
BA  & 52.95  & 50.71  & 50.07  & 87.29 & \textbf{100.00} & 87.69 & \textbf{100.00} & 91.58 & 98.33 \\
Std & 2.13   & 0.87   & 0.15   & 16.89 & 0.00   & 20.64 & 0.00   & 17.42 & 8.98  \\
FP  & 94.11  & 98.58  & 99.85  & 11.67 & 0.00   & 0.00  & 0.00   & 0.00  & 0.00  \\
Std & 4.25   & 1.75   & 0.29   & 11.78 & 0.00   & 0.00  & 0.00   & 0.00  & 0.00  \\
\midrule
\multicolumn{10}{c}{Operating mode 4: Lower Cruising} \\
\midrule
BA  & 50.00  & 50.00  & 50.00  & 93.33 & \textbf{100.00} & 88.33 & \textbf{100.00} & 91.96 & 98.33 \\
Std & 0.00   & 0.00   & 0.00   & 17.00 & 0.00   & 21.15 & 0.00   & 17.99 & 8.98  \\
FP  & 100.00 & 100.00 & 100.00 & 0.00  & 0.00   & 0.00  & 0.00   & 0.00  & 0.00  \\
Std & 0.00   & 0.00   & 0.00   & 0.00  & 0.00   & 0.00  & 0.00   & 0.00  & 0.00  \\
\midrule
\multicolumn{10}{c}{Operating mode 5: Landing} \\
\midrule
BA  & 50.00  & 50.00  & 50.00  & 93.33 & \textbf{99.43}  & 88.33 & 99.04  & 91.72 & 97.08 \\
Std & 0.00   & 0.00   & 0.00   & 17.00 & 0.80   & 21.15 & 1.79   & 18.51 & 9.09  \\
FP  & 100.00 & 100.00 & 100.00 & 0.00  & 1.13   & 0.00  & 1.92   & 0.00  & 2.50  \\
Std & 0.00   & 0.00   & 0.00   & 0.00  & 1.60   & 0.00  & 3.59   & 0.00  & 5.01 \\\bottomrule
\multicolumn{10}{l}{M: HELM trained with Mixed dataset (Source + Target)}\\
\multicolumn{10}{l}{A: Results using ADAU}\\
\end{tabular}
\label{tb:AGTF}
\end{table}
\endgroup

\textbf{Results.} %
The results under each operating mode are presented in Table~\ref{tb:AGTF} and illustrated in Figure~\ref{fig:AGTF30Results}. As expected, HELM trained solely on samples from the second operating mode of the target detects all other operating modes as anomalous, which leads to 100\% FPR and thus a BA of 50\%. For these modes, adding data from the source can only help, but the improvement is limited by the distribution shift between the two planes. In these cases, performing the alignment significantly improves all results, with the BA reaching up to 100\% as long as the number of samples added from the source remains within the same order of magnitude as the number of training samples from the target (up to a factor 2-3 as for the CWRU case study).

All tested models give output over all faults and conditions that are statistically different from each others with $p$-values below $10^{-4}$. The only exception being the ADAU model trained with 5 and 10\% of the source dataset. This is expected since their BA are almost everywhere 100\%. At the global level, the ADAU approach gives outputs significantly different to the mixed approach with a $p$-value below $10^{-6}$.

    \subsection{MNIST - MNIST-M as Anomaly Detection}
Complex industrial systems often operate in an extremely stable manner as long as the operating conditions are not changed either due to a new operation (\eg a flight transitioning from cruising to landing) or an environmental change (\eg seasonality).

\begin{figure}
\centering
\includegraphics[width=5cm]{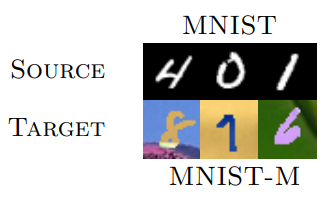}
\caption{\textbf{MNIST - MNIST-M.} Example of samples taken from both MNIST and MNIST-M datasets. Figure taken from~\cite{ganin2016domain}}
\label{fig:MNISTexamples}
\end{figure}

\textbf{Experiment.} %
To test the proposed solution in a context with more diversity within the main class, and to show the generalisability of the solutions to other fields, we propose in this last experiment to perform the alignment on the MNIST to MNIST-M transfer task~\cite{ganin2016domain}. The MNIST dataset~\cite{lecun1998gradient} contains 55\,000 handwritten images of the 10 digits. The images are of size 28$\times$28$\times$3. The MNIST-M dataset contains patches from the BSDS500~\cite{arbelaez2010contour}, clipped with MNIST digits, as illustrated in Figure~\ref{fig:MNISTexamples}. 

Here we consider one digit as the main class (digit 0), and all others as anomalies. The dataset is made up of 5\,455 images of class 0 for source and target and 49\,545 anomalies. The proposed loss in Equation~\eqref{eq:MDSLoss} necessitates a distance computation between samples. A relevant distance measure between images is still an open research question. In this work we choose to use the image Euclidean distance (IMED)~\cite{wang2005,nakhmani2013new} for three reasons. First, it is similar to the traditional Euclidean distance used in the other case studies. Second, experiments have shown its robustness to scale, translation, rotation, image size, and noise~\cite{nakhmani2013new}. Third, since its implementation requires only the pixel-wise Euclidean distance computation after the application of a Gaussian blur kernel to the image, it can be computed on the vectorised representation of the image. This allows us to use the same architecture as in other experiments. 
If $N^2$ is the size of the images, the IMED distance is defined as:
\begin{equation}
d_{\mathrm{IMED}} = \sqrt{\sum_{i=1}^{N^2} \sum_{j=1}^{N^2} g_{ij}\cdot (x_i - y_i)\cdot (x_j-y_j)},
    \label{eq:imed}
\end{equation}
where
\begin{equation}
g_{ij} = \frac{1}{2\pi\sigma^2}\exp{\left( \frac{-dist(P_i,Pj)^2}{2\sigma^2}\right)}
\end{equation}
and where $dist(P_i,Pj)$ denotes the Euclidean distance between the pixels $i$ and $j$.

Similar to other experiments, we infer the number of neurons in our architecture with the elbow approach on HELM. The HELM auto-encoder has 16 neurons and, therefore, the two layers of the feature extractor also have 16 neurons. The one-class classifier has 100 neurons. 

\begin{figure}
\centering
\includegraphics[width=\imwidth]{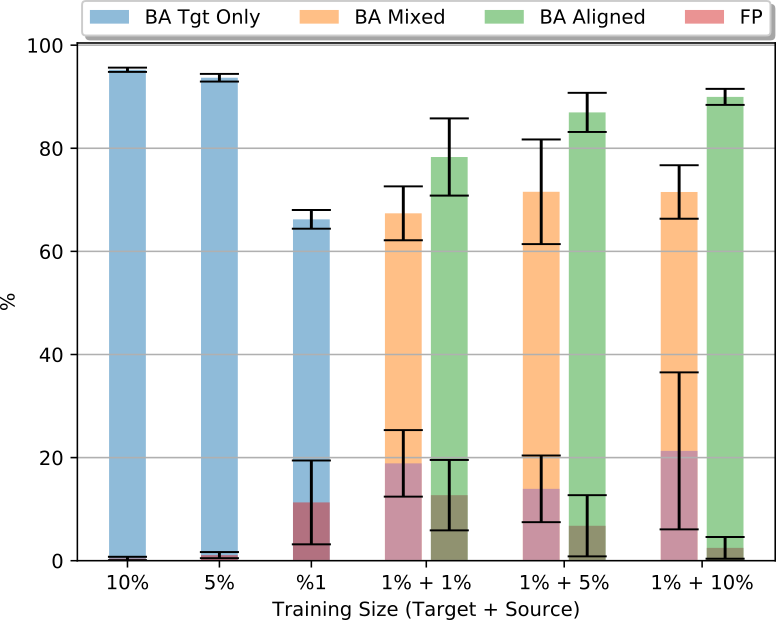}
\caption{\textbf{Anomaly Detection Transfer on the MNIST to MNIST-M task.} The healthy class is the digit 0, while all other digits are considered anomalies.}
\label{fig:MNIST}
\end{figure}

\begingroup
\setlength{\tabcolsep}{4pt}
\begin{table}[]
\centering
\caption{MNIST - Results (BA, FP, and their standard deviation Std) for the MNIST to MNIST-M transfer task.}
\begin{tabular}{lrrr|rr|rr|rr}
\toprule
& \multicolumn{3}{c|}{Target Only} & \multicolumn{6}{c}{1\% Target + Source}\\
& \multicolumn{3}{c|}{HELM} & \multicolumn{1}{c}{M} & \multicolumn{1}{c|}{A} & \multicolumn{1}{c}{M} & \multicolumn{1}{c|}{A} & \multicolumn{1}{c}{M} & \multicolumn{1}{c}{A}\\
& 10\% & 5\% & 1\% & \multicolumn{2}{c|}{+1\%} & \multicolumn{2}{c|}{+5\%} & \multicolumn{2}{c}{+10\%}\\
\midrule
BA  & 95.23 & 93.69 & 66.21 & 67.38 & 78.31 & 71.56 & 86.96 & 71.52 & \textbf{89.97} \\
Std & 0.41  & 0.74  & 1.82  & 5.22  & 7.49  & 10.15 & 3.80  & 5.19  & 1.56  \\
FP  & 0.43  & 1.08  & 11.31 & 18.88 & 12.71 & 13.94 & 6.78  & 21.31 & 2.49  \\
Std & 0.33  & 0.59  & 8.13  & 6.45  & 6.83  & 6.46  & 5.94  & 15.23 & 2.11  \\ \bottomrule
\multicolumn{10}{l}{M: HELM trained with Mixed dataset (Source + Target)}\\
\multicolumn{10}{l}{A: Results using ADAU}\\
\end{tabular}
\label{tb:MNIST}
\end{table}
\endgroup

\textbf{Results.} %
The results are presented in Table~\ref{tb:MNIST} and illustrated in Figure~\ref{fig:MNIST}. The results show that when the number of training examples from the target diminishes, the one-class classifier loses accuracy from 95\% when trained with 10\% of the main class to 66.21\% when trained with 1\%. This demonstrates the inability of the model to learn the main class properly, and indicates a probable large variability in the main class. In this case, simply mixing the data without alignment helps on average, but with a high results variability. This apparent instability in the results is probably due to the random selection of the source and target samples used for training. With the alignment, results improve significantly, reaching up to 90\% BA once 10\% of the source dataset is added.

All model outputs are statistically different from each other's with $p$-values below $10^{-6}$ according to both the GLM and McNemar tests. Also at the global level, when all models using ADAU are compared to all models using mixed source and target data, the models' difference is significant with $p$-values below $10^{-6}$.




\section{Conclusion}
Throughout the three presented case studies, we demonstrated that the accuracy of data-driven anomaly detection methods depends on whether the training data covers the variability of the main class. When this is not the case, data from other related sources can be used in complement. Yet when the different data sources have a domain shift (e.g. a different engine setup, images of a different nature, or bearings with a different load), the anomaly detector can fail to recognise the main class. This can be mitigated by aligning the domains with an adversarial architecture while conserving the inter-sample relationships with the proposed loss. This setup has two limits, however. First, the source and target domain should have a number of training samples in the same order of magnitude, probably due to the difficulty of training the domain discriminator in an unbalanced setup. This might limit the additional experience that can be introduced by the source. Second, this approach benefits the final results if the source offers additional unobserved operating conditions. In stable operation, similar to the CWRU experiments, results showed little difference between an AD trained on the few available target samples and an AD trained on both source and target.
In the future, this work may be extended to multi-source alignment in an attempt to collect operating condition experience from whole fleets of units.

\section*{Acknowledgment}
    \aknow

    \bibliography{references}

\end{document}